\documentclass[preprint,10pt,comma]{elsarticle}
\usepackage{geometry}
\usepackage{import}
\usepackage{enumitem}
\usepackage{amsfonts}
\usepackage{algorithmic}
\usepackage{graphicx}
\usepackage{adjustbox}
\usepackage{changepage}
\usepackage{amsmath} % assumes amsmath package installed
\usepackage{amssymb}  % assumes amsmath package installed
\usepackage{array}
\usepackage{lipsum}
\usepackage{subcaption}
\usepackage{multirow}
\usepackage{hhline}
\usepackage{hyperref}
\usepackage{booktabs}    % For professional-quality tables (\toprule, \midrule, \bottomrule)
\usepackage{tabularx}    % For tables with columns that automatically adjust to fill a specified width
\usepackage{textcomp}
\usepackage{amsthm}
\usepackage{stfloats}
\usepackage{float}
\usepackage{verbatim}
\usepackage{balance}
\usepackage{cleveref}
\usepackage{setspace}
\usepackage{xcolor}
\usepackage{tikz}
\usepackage{bbm}

\begin{document}

\begin{frontmatter}
    \title{Performance Decay in Deepfake Detection: The Limitations of Training on Outdated Data}
    
    \author[ATI]{Jack Richings\corref{cor1}}
    \ead{jrichings@turing.ac.uk}
    
    \author[ATI]{Margaux Leblanc}
    \author[ATI]{Ian Groves}
    \author[ATI]{Victoria Nockles}

    \cortext[cor1]{Corresponding author}
    \address[ATI]{Defence AI Research (DARe), The Alan Turing Institute, London NW1 2DB, UK.}

    \begin{abstract}
    The continually advancing quality of deepfake technology exacerbates the threats of disinformation, fraud, and harassment by making maliciously-generated synthetic content increasingly difficult to distinguish from reality. We introduce a simple yet effective two-stage detection method that achieves an AUROC of over 99.8\% on contemporary deepfakes. However, this high performance is short-lived. We show that models trained on this data suffer a recall drop of over 30\% when evaluated on deepfakes created with generation techniques from just six months later, demonstrating significant decay as threats evolve. Our analysis reveals two key insights for robust detection. Firstly, continued performance requires the ongoing curation of large, diverse datasets. Second, predictive power comes primarily from static, frame-level artifacts, not temporal inconsistencies. The future of effective deepfake detection therefore depends on rapid data collection and the development of advanced frame-level feature detectors.
    \end{abstract}

    \begin{keyword}
    Deepfake detection, Performance decay, Concept drift, Media forensics, Disinformation
    \end{keyword}

\end{frontmatter}

% The main body of your article starts here

% \begingroup\renewcommand\thefootnote{}
% \footnotetext{\textit{
% This work was supported by the Turing's Defence and Security programme through a partnership with the UK government in accordance to the framework agreement between Dstl \& The Alan Turing Institute.}}
% \endgroup

\section{INTRODUCTION}

Deepfakes are AI-generated media that digitally alter or superimpose a person's likeness onto another individual, convincingly making them appear to say or do something they never did. Deepfakes presents a multifaceted and escalating threat to individuals, society, and democratic processes. The malicious application of this technology causes many harms, including severe individual and reputational damage through non-consensual pornography and cyberbullying, sophisticated financial fraud via voice cloning and impersonation of executives, and significant political disruption by spreading disinformation and manipulating public opinion \cite{citron2019deep, kreps2023how}. Furthermore, the proliferation of deepfakes contributes to a broader societal erosion of trust in digital media, a phenomenon known as the "liar's dividend" \cite{Schiff2023} which undermines the concept of objective truth and presents major psychological, legal, and ethical challenges in a world where seeing is no longer believing.

\noindent \textbf{Related work.} The fields of deepfake generation and detection are locked in an arms race, with the rapid improvements in deepfake realism spurring the creation of new detection methods. Early methods focused on identifying specific artifacts inherent to the generation process, such as visual inconsistencies at blending boundaries, geometric warping distortions, and unnatural facial expressions. Researchers focussed on analysing biological and physical signals, scrutinising inconsistencies in behaviours like eye blinking patterns, mouth movements, and posture to identify deepfakes~\cite{li2018ictuoculiexposingai, matern2019, nguyen2020eyebrows, haliassos2021lipsdontliegeneralisable}.

Data-driven approaches are now the standard. Approaches include supervised methods (on a frame-by-frame basis \cite{rossler2019faceforensicslearningdetectmanipulated} or in a spatio-temporal approach \cite{zheng2021exploringtemporalcoherencegeneral}) to self-supervised anomaly detection methods \cite{haliassos2022leveragingrealtalkingfaces, cai2023marlinmaskedautoencoderfacial}. Both Convolutional Neural Networks (CNNs) and, more recently, Vision Transformers (ViTs) are commonly employed architectures.
% Architectures based on Convolutional Neural Networks (CNNs) formed the backbone of early deep learning-based detectors, primarily trained to identify spatial artifacts left by Generative Adversarial Networks (GANs).
However, the landscape has shifted with the rise of diffusion models, which produce higher-fidelity fakes with different, more subtle artifacts. Recent surveys have found detection methods designed for content created using Generative Adversarial Networks (GANs) often fail to generalise to diffusion-generated deepfakes \cite{xu2025recentadvancesgeneralizablediffusiongenerated}, motivating both the development of larger, more complex models and alternative approaches.

To address the increasing sophistication of deepfakes, two major trends have emerged: multi-modal analysis and proactive detection. Multi-modal approaches leverage inconsistencies across different data streams. Recent work includes methods that learn audio-visual consistency~\citep{Miao_Guo_Liu_Wang_2025, haliassos2022leveragingrealtalkingfaces} and the exploration of Multi-modal Large Language Models like GPT-4V as zero-shot deepfake detectors~\citep{jia2024chatgptdetectdeepfakesstudy, shahzad2024goodchatgptaudiovisualdeepfake}. Proactive detection, in contrast, aims to prevent or disrupt forgery by embedding information into media pieces before distribution. Techniques range from embedding adversarial perturbations to sophisticated, learnable, watermarks that are robust to benign transformations but which can potentially be destroyed by malicious manipulation~\citep{li2025bigbrotherwatchingproactive, zhao2025sokwatermarkingaigeneratedcontent}.

Both multi-modal analysis and watermarking face serious challenges as mitigation strategies. Multi-modal analysis, which seeks inconsistencies between audio and video streams, is likely to become less effective over time. As generative methods improve, they will perfect the synchronisation of facial and mouth movements with synthesised audio. This mirrors the trajectory of earlier detection techniques that focused on simple physical artifacts like unnatural eye blinking, which have now been largely overcome by more advanced models. Furthermore, multi-modal analysis faces challenges when dealing with real-world footage that is often of poor quality. Blurring, low frame rates, compression artifacts, and low-fidelity audio can obscure the very inconsistencies these detectors are designed to find, leading to a high rate of failure.
Watermarking, while promising, is not a wholesale solution. Its efficacy depends on widespread, near-universal adoption, an unrealistic assumption. Malicious actors with significant resources can train their powerful generative models on vast archives of data scraped from the internet. Well-intentioned but careless developers may fail to implement sufficiently robust watermarking schemes. Additionally, watermarks and other countermeasures will become the targets of adversarial attacks. Sophisticated adversaries could develop deep learning-based methods to specifically remove, degrade or imitate embedded watermarks, making the protection ineffectual. Other provenance tracking methods, for example those based on blockchain technology~\citep{NewsProvenanceProject, FactomWhitepaper}, face significant technical hurdles and are unlikely to see adoption in the near future except for specific use cases.

In summary, all attempts to mitigate the harm caused by deepfake content are engaged in an arms race, with no approach offering a universal and definite solution. In this fast-paced context, it is important to assess the rate at which deepfake detection models become redundant, and to minimise the time required in training new models to counteract the latest deepfakes.

\noindent \textbf{Contributions.} We make several contributions. Firstly, we introduce a two-stage pipeline that achieves strong performance on DeepSpeak version 1.1 and version 2.0. The strong performance of this relatively simple model demonstrates that even cutting-edge deepfakes can be detected with simple frameworks if the training dataset is good enough. Secondly, we evaluate the contribution of the model's components, providing guidance on where effort is best spent in model optimisation. Thirdly, we use the models trained using our pipeline to estimate drop in performance attributable to six months of advancement in deepfake generation. Finally, we estimate the impact that dataset size (individual diversity) has on model performance, informing strategies for the data-efficient development of future models and the curation of datasets.

\noindent \textbf{Paper structure.} In section~\ref{sec:data} we justify our focus on the DeepSpeak dataset, and review its key characteristics. In section~\ref{sec:methods} we describe a two-stage pipeline for deepfake video detection, as well as the details of our fine-tuning methods. Section~\ref{sec:results} details the results of our experiments, including the overall performance of the pipeline, the contribution of its components and the results of generalisation-testing. In section~\ref{sec:discussion} we discuss the implications of the results presented in section~\ref{sec:results} and comment on the ability of models to detect new kinds of deepfake. We conclude in section~\ref{sec:conclusion}.

%%%%%%%%%%%%%%%%%%%%%%%%%%%%%%%%%%%%%%%%%%%%%%%%%%%%%%%%%%%%%%%%%%%%%%%%%%%%%%%%
\section{DATA}
\label{sec:data}

\subsection{The DeepSpeak dataset}
The DeepSpeak dataset~\citep{Barrington2024DeepSpeak} consists of real and deepfake footage of people talking and gesturing in front of webcams. The DeepSpeak dataset distinguishes itself from other forensic-themed public datasets by its comprehensive nature, currency, and continuous updates. While datasets like FaceForensics~\citep{rossler2018faceforensicslargescalevideodataset} and DFDC~\citep{dolhansky2020deepfakedetectionchallengedfdc}  are significant, DeepSpeak's commitment to regular updates with emerging deepfake technologies makes it particularly relevant for current research. The focus on diverse individuals and the explicit consent further enhance its utility and ethical standing.

There are two versions of the DeepSpeak dataset, version 1.1 and version 2.0, which contain videos of different people and use an overlapping but different set of techniques for manipulating video and audio. Version 1.0 was released in August 2024, and version 2.0 was generated April 2025. A full breakdown of the tools and techniques used to generate the datasets is given in \citep{Barrington2024DeepSpeak}, but we summarise the key properties for completeness.

The DeepSpeak dataset was generated using a range of identity-cloning approaches. \textit{Face-swap} deepfakes replace the face (chin to eyebrow, cheek to cheek) of the source individual with the face of a target individual. Source and target individuals are matched using the similarity of their CLIP~\citep{radford2021learningtransferablevisualmodels} embeddings to maximise the realism of the deepfake. \textit{Lip-sync} deepfakes modify only the mouth region, to make it consistent with a new audio track. \textit{Avatar} deepfakes create a video from a single static image. Only version 2.0 of the DeepSpeak dataset contains avatar deepfakes. Each kind of deepfake generation is implemented via several different engines. For example, avatar deepfakes are created using the LivePortrait~\citep{liveportrait}, HelloMeme~\citep{hellomeme} and Memo~\citep{memo} engines. The number of each kind of deepfake in each dataset, as well as the specific engines used to create each kind of deepfake, is listed in Table~\ref{tab:deepspeak_counts}.

\begin{table*}[h]
\centering
\caption{Summary of the makeup of the DeepSpeak v1.1 and v2.0: share of fake videos and their generative engines in train and test sets.}
\label{tab:deepspeak_counts}

% --- Subtable for DeepSpeak v1.1 ---
\begin{subtable}{\textwidth}
    \centering
    \caption{Summary of the makeup of DeepSpeak v1.1}
    \begin{tabular}{@{}llcc@{}}
        \toprule
        \textbf{Deepfake Technique} & \textbf{Deepfake Engine} & \textbf{Train Set} (Share) & \textbf{Test Set} (Share) \\
        \midrule
        \multirow{2}{*}{Lip-Sync}
        & Wav2Lip ~\citep{Prajwal2020W2L} & 1,397 (26\%) & 399 (27\%) \\
        \cmidrule(l){2-4}
        & VideoRetalking ~\citep{Cheng2022VideoReTalking} & 1,484 (28\%) & 412 (27\%) \\
        \midrule
        \multirow{3}{*}{Face Swap}
        & FaceFusion ~\citep{FaceFusionGithub} & 797 (15\%) & 228 (15\%) \\
        \cmidrule(l){2-4}
        & FaceFusion ~\citep{FaceFusionGithub} + CodeFormer GAN ~\citep{Zhou2022CodeFormerGAN} & 796 (15\%) & 228 (15\%) \\
        \cmidrule(l){2-4}
        & FaceFusion Live ~\citep{FaceFusionGithub} & 825 (16\%) & 230 (15\%) \\
        \midrule
        \multicolumn{2}{@{}l}{\textbf{Total number of Fake videos per set}} & \textbf{5,299} & \textbf{1,497} \\
        \multicolumn{2}{@{}l}{\textbf{Total number of Real videos per set}} & \textbf{5,251} & \textbf{1,416} \\
        \bottomrule
    \end{tabular}
\end{subtable}

\vspace{0.5cm} % Space between the two sub-tables

% --- Subtable for DeepSpeak v2.0 ---
\begin{subtable}{\textwidth}
    \centering
    \caption{Summary of the makeup of DeepSpeak v2.0}
    \begin{tabular}{@{}llcc@{}}
        \toprule
        \textbf{Deepfake Technique} & \textbf{Deepfake Engine} & \textbf{Train Set} (Share) & \textbf{Test Set} (Share) \\
        \midrule
        \multirow{2}{*}{Lip-Sync}
        & Diff2Lip ~\citep{mukhopadhyay2024diff2lip} & 848 (15\%) & 192 (14\%) \\
        \cmidrule(l){2-4}
        & LatentSync ~\citep{li2024latentsync} & 1,099 (19\%) & 277 (20\%) \\
        \midrule
        \multirow{4}{*}{Face Swap}
        & INSwapper ~\citep{inswapper} & 205 (4\%) & 57 (4\%) \\
        \cmidrule(l){2-4}
        & INSwapper ~\citep{inswapper} + CodeFormer ~\citep{Zhou2022CodeFormerGAN}& 215 (4\%) & 53 (4\%) \\
        \cmidrule(l){2-4}
        & SimSwap ~\citep{chen2020simswap} & 211 (4\%) & 60 (4\%) \\
        \cmidrule(l){2-4}
        & SimSwap ~\citep{chen2020simswap} + RestoreFormer ~\citep{wang2023restoreformer} & 228 (4\%) & 48 (3\%) \\
        \midrule
        \multirow{3}{*}{Avatar}
        & LivePortrait ~\citep{liveportrait} & 1,002 (17\%) & 252 (18\%) \\
        \cmidrule(l){2-4}
        & HelloMeme ~\citep{hellomeme} & 978 (17\%) & 239 (17\%) \\
        \cmidrule(l){2-4}
        & Memo ~\citep{memo} & 1,007 (17\%) & 238 (17\%) \\
        \midrule
        \multicolumn{2}{@{}l}{\textbf{Total number of Fake videos per set}} & \textbf{5,793} & \textbf{1,416} \\
        \multicolumn{2}{@{}l}{\textbf{Total number of Real videos per set}} & \textbf{7,513} & \textbf{1,863} \\
        \bottomrule
    \end{tabular}
\end{subtable}
\end{table*}

\subsection{Individual identities}
\label{sec:identities}
Each version of the DeepSpeak datasets contains well over ten thousand videos, but these videos contain approximately two hundred unique identities. In other words, each person in the dataset appears in dozens of videos, split roughly evenly between genuine and deepfake videos. The train and test sets are split at the individual level, meaning that each unique individual appears in either the train or the test set, but not both. However, there is no specific validation set which maintains this split.
To prevent overfitting to individual identities, construction of a validation set is a necessary step. We generate our validation set by randomly splitting the list of individual identities in the train dataset into a train list and a validation list. The full validation dataset is obtained by taking all videos containing individuals in the validation list.

When generating subsets of the dataset for experiments related to dataset size (see Section~\ref{sec:methods_finetuning}) we split the dataset at the individual level. For example, to construct a dataset 50\% the size of the full training dataset, we randomly select 50\% of the individuals in the training set, and use all videos of those individuals as the subset.

\section{METHODS}
\label{sec:methods}
\begin{figure*}[!htbp] % figure* makes it span two columns
    \centering
    \includegraphics[width=0.9\textwidth]{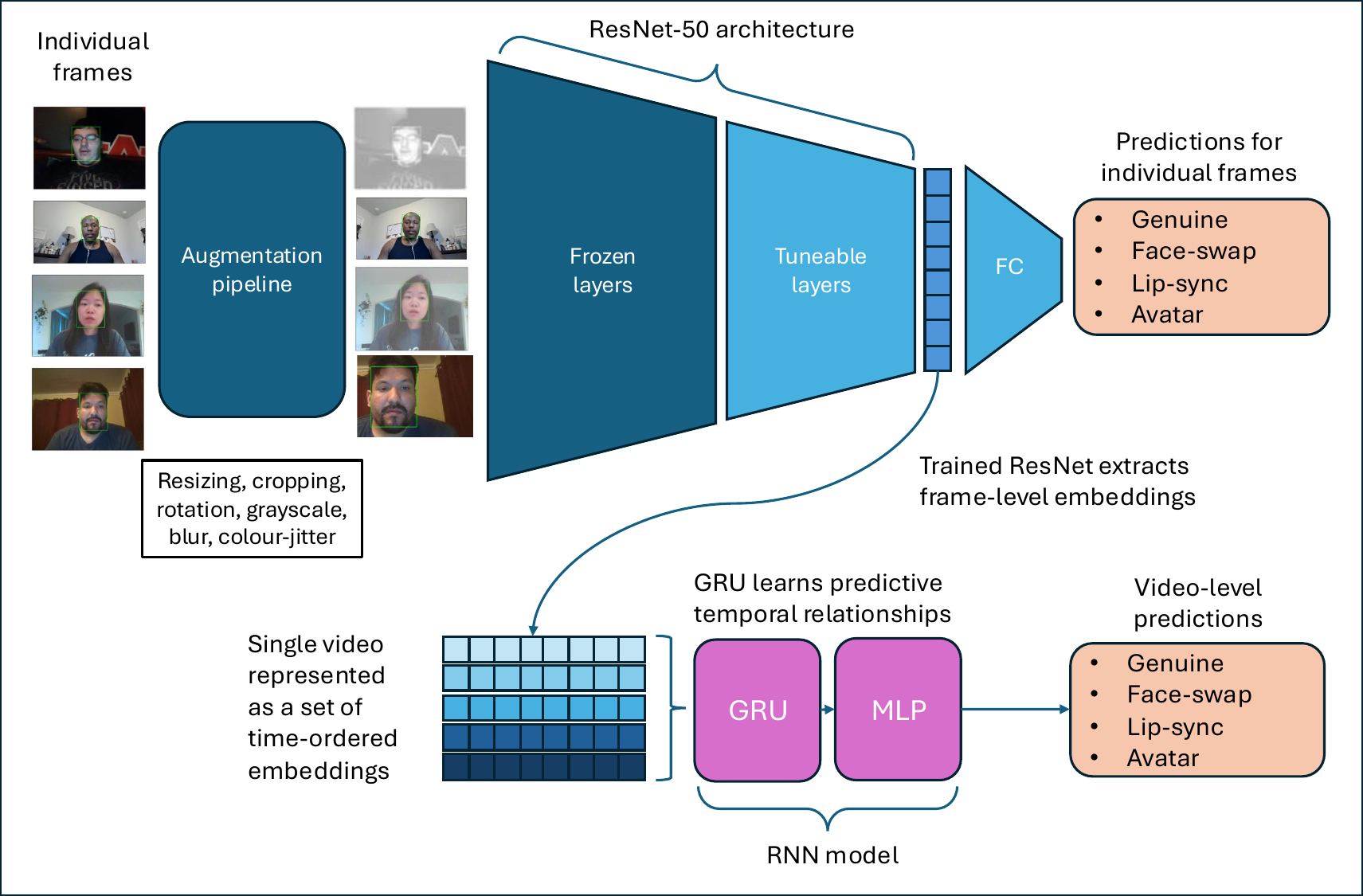}
    \caption{Overview of the two-stage ResNet-RNN model architecture and training process. Individual video frames are first processed through an augmentation pipeline. A pre-trained ResNet-50, with frozen early layers, learns frame-level embeddings before predicting frame-level labels via the fully-connected (FC) layer. Learned embeddings are then collated and time-ordered for each video and then fed into a GRU-based RNN model, which learns temporal relationships and outputs video-level predictions.}
    \label{fig:resnet_rnn_layout}
\end{figure*}

We train a deepfake video detection model using a two-step approach, illustrated in Fig. ~\ref{fig:resnet_rnn_layout}.
\begin{enumerate}
    \item Train a Convolutional Neural Network (CNN) to classify individual frames of videos.
    \item Train a Recurrent Neural Network (RNN) to classify time-ordered sets of features extracted from frames using the CNN trained in step 1.
\end{enumerate}
As the CNN we use is based on the ResNet-50 architecture~\citep{he2015deepresiduallearningimage}, we refer to the final model as a ResNet-RNN.

\subsection{CNN training}
\label{sec:cnn_training}
To construct the dataset for training and evaluating the CNN model, we use a pre-trained Multi-Task Cascaded Convolutional Network (MTCNN) model~\citep{Zhang_2016, timesler_facenet-pytorch_2019}, to perform face detection. We set the side length of the bounding box to be 320 pixels, with a margin of 16 pixels around the edge of the face. We perform face extraction on 5 frames per second of video, sampled uniformly in time, up to a maximum of 50 frames. Temporal downsampling sampling is employed to mitigate effects of varying video frame rate on model performance. Our train and test sets consist of 380,000 and 106,000 images respectively.

Our CNN model uses a standard ResNet-50 architecture~\citep{he2015deepresiduallearningimage}. The ResNet-50 model consists of 4 blocks, each containing 3 convolutional layers. We initialise the model using ImageNet weights~\citep{russakovsky2015imagenetlargescalevisual}. We freeze the first two blocks during training, as these layers encode general low-level features which transfer well between tasks. We train the model using the Adam optimiser~\citep{kingma2017adammethodstochasticoptimization} with a learning rate of $10^{-4}$, an L2 norm of $10^{-5}$, and an early stopping based on the validation loss, with a patience of 10 epochs. During training, we employ an aggressive augmentation pipeline to alleviate overfitting. Our augmentation pipeline consists of random combinations of horizontal flipping, rotation, resizing, cropping, colour-jittering and Gaussian blurring.

When evaluating this model, we take predictions from individual frames of a video and combine them by computing the mean of the softmax probabilities of each frame to produce a single video-level prediction.

\subsection{RNN training}
\label{sec:methods_rnn}
The RNN model consists of a set of Gated Recurrent Unit (GRU) layers~\citep{chung2014empiricalevaluationgatedrecurrent}, followed by a set of fully-connected layers. The fully connected layers use ReLU activations, and we employ dropout in both the fully-connected and GRU layers to mitigate overfitting. The results presented in this paper were obtained using a network with 5 bidirectional GRU layers with a hidden dimension of 512, and 4 fully connected layers.

We use the trained CNN to extract embeddings from the image dataset, taking the activations from the final layer before the classification head. The embeddings have a feature dimension of 2048. All embeddings for a given video are arranged chronologically to form a 2D representation of the video.

To standardise video duration, we use cropping and padding along the time axis, incorporating randomisation during training to prevent overfitting. For videos longer than the target length, we randomly crop a section during training. For testing, we always crop from the beginning to ensure consistent results. If a video is shorter than the target length, we pad the beginning and end of the video by repeating the first and last frames respectively. For testing, we only repeat the last frame. This dynamic cropping and padding for each training epoch exposes the model to different parts of the video, reducing overfitting. To further prevent overfitting, we mask a random subset of features (10\%) in each video sample. The selected features remain hidden across all frames within that video, however different features are masked for each video at each epoch.

During training we employ five-fold cross validation, with an early stopping criterion based on the validation loss with a patience of 10 epochs. The training and validation sets in each split are constructed such that there is no overlap in identities between the sets. All five trained models are evaluated on the test set to obtain an average performance.

We train the models using the Adam optimiser~\citep{kingma2017adammethodstochasticoptimization}, with a learning rate of $10^{-4}$ and an L2 norm of $10^{-5}$. We experimented with class-weighted loss functions and a learning-rate scheduler, but found they did not improve performance.

\subsection{Fine-tuning}
\label{sec:methods_finetuning}
A key question which we aim to answer is \textit{``Is it possible to fine-tune an old deepfake detection model with limited new data? When does training a new model from scratch become the preferred option?''}. To explore this question, we fine-tune a ResNet model trained on DeepSpeak version 1.1 on incrementally larger fractions of the DeepSpeak version 2.0 dataset. Subsets of version 2.0 are constructed using the approach described in section~\ref{sec:identities}. All layers of the model trained on the version 1.1 dataset are frozen, except for the fourth block and the fully connected layers. The training procedure is otherwise identical to the approach described in section~\ref{sec:cnn_training}, apart from the learning rate, which we set to $2\times 10^{-5}$. Since the large majority of predictive power arises from the ResNet component our ResNet-RNN model, we present fine-tuning results only for the ResNet component. This avoids the need to disentangle components of the network, whilst still providing valuable information.

\subsection{FaceNet embeddings}
To assess the contributions of the CNN and RNN components of our pipeline, we perform feature extraction using a different feature extraction model. We use the FaceNet model~\citep{Schroff_2015, timesler_facenet-pytorch_2019}, an Inception-ResNet~\citep{szegedy2015rethinkinginceptionarchitecturecomputer} trained on the VGGFace2 dataset~\citep{vggface2}\footnote{It is reasonable to ask why we do not use ImageNet features as a comparator, or why we do not initialise weights using FaceNet. FaceNet weights likely offer superior out-of-the-box performance, but the architecture is limited to Inception-ResNet. ImageNet weights are available for many architectures, and so by initialising from ImageNet weights we make it easier to compare new training methods and architectures to our approach.}. Crucially, the FaceNet model was trained using only real images, and therefore its embeddings do not capture features which occur only in deepfake images. The training procedure for the RNN trained using FaceNet embeddings is identical to the procedure used to train the RNN using deepfake-specific embeddings (Section~\ref{sec:methods_rnn}).

%%%%%%%%%%%%%%%%%%%%%%%%%%%%%%%%%%%%%%%%%%%%%%%%%%%%%%%%%%%%%%%%%%%%%%%%%%%%%%%%
\section{RESULTS}

\label{sec:results}

This section reports the test set evaluation of the ResNet-RNN model and its constituent components, detailed in Section~\ref{sec:methods}, on the DeepSpeak dataset.

\subsection{Impact of novel deepfake generation techniques}
We show that the ResNet-RNN model is capable of distinguishing real and deepfake videos with both high recall and precision. Table~\ref{tab:cross_evaluation_results} details the performance of the ResNet-RNN model on both versions 1.1 and 2.0 of the DeepSpeak dataset. For both versions, we achieve an AUROC in excess of 99\%. The implication is that comparatively simple methods are still capable of detecting cutting edge deepfakes given data of sufficient quality and quantity.

\begin{table*}[h]
\centering
\caption{Cross-evaluation of the ResNet-RNN models trained on different dataset versions. We report overall Accuracy and AUROC, alongside per-class Precision, Recall, and F1-Score for 'real' and 'fake' classes. The results show high performance when training and testing on matched versions (v1/v1, v2/v2) but a significant performance drop in cross-version evaluation (v1/v2, v2/v1), highlighting a dataset shift. All metrics are presented as mean\% $\pm$ std. dev\% over multiple runs.}
\label{tab:cross_evaluation_results}
% Using resizebox to ensure the table fits within the text width of a two-column layout
% \resizebox{\textwidth}{!}{%
\begin{tabular}{@{}llcc l ccc@{}}
\toprule
\multicolumn{2}{c}{\textbf{Dataset}} & & & & \multicolumn{3}{c}{\textbf{Per-Class Metrics}} \\
\cmidrule(r){1-2} \cmidrule(l){6-8}
\textbf{Train} & \textbf{Test} & \textbf{Accuracy (\%)} & \textbf{AUROC (\%)} & \textbf{Class} & \textbf{Precision (\%)} & \textbf{Recall (\%)} & \textbf{F1-Score (\%)} \\
\midrule

% --- Train v1, Test v1 ---
\multirow{2}{*}{V1.1} & \multirow{2}{*}{V1.1} & \multirow{2}{*}{$98.06 \pm 0.45$} & \multirow{2}{*}{$99.67 \pm 0.03$} & Real & $98.66 \pm 0.19$ & $97.34 \pm 0.98$ & $97.99 \pm 0.48$ \\
& & & & Fake & $97.53 \pm 0.90$ & $98.74 \pm 0.18$ & $98.13 \pm 0.43$ \\
\midrule

% --- Train v2, Test v2 ---
\multirow{2}{*}{V2.0} & \multirow{2}{*}{V2.0} & \multirow{2}{*}{$99.30 \pm 0.05$} & \multirow{2}{*}{$99.82 \pm 0.09$} & Real & $96.25 \pm 0.31$ & $99.72 \pm 0.26$ & $97.95 \pm 0.15$ \\
& & & & Fake & $99.94 \pm 0.05$ & $99.21 \pm 0.07$ & $99.57 \pm 0.03$ \\
\midrule

% --- Train v1, Test v2 ---
\multirow{2}{*}{V1.1} & \multirow{2}{*}{V2.0} & \multirow{2}{*}{$69.53 \pm 2.80$} & \multirow{2}{*}{$90.18 \pm 1.08$} & Real & $35.36 \pm 2.06$ & $95.76 \pm 0.71$ & $51.61 \pm 2.24$ \\
& & & & Fake & $98.67 \pm 0.23$ & $64.19 \pm 3.37$ & $77.73 \pm 2.55$ \\
\midrule

% --- Train v2, Test v1 ---
\multirow{2}{*}{V2.0} & \multirow{2}{*}{V1.1} & \multirow{2}{*}{$87.09 \pm 1.55$} & \multirow{2}{*}{$94.50 \pm 1.33$} & Real & $96.99 \pm 0.47$ & $75.81 \pm 3.34$ & $85.06 \pm 2.04$ \\
& & & & Fake & $81.09 \pm 2.16$ & $97.77 \pm 0.39$ & $88.63 \pm 1.21$ \\
\bottomrule
\end{tabular}%
% }
\end{table*}

Table~\ref{tab:cross_evaluation_results} also details the performance of the model trained on the DeepSpeak version 2.0 dataset when evaluated on the test set of DeepSpeak version 1.1, and vice versa. This evaluation provides a valuable insight into the expected drop in performance associated with advancing deepfake technology on a timescale of months. There is no overlap in the individuals appearing in the two test sets, nor is there any overlap in the exact engines used to produce the deepfakes \footnote{DeepSpeak version 1.1 and 2.0 both use FaceFusion to generate face-swap deepfakes, however they use different face-swap generators within the FaceFusion library.}. By evaluating version 2.0 on version 1.1, we obtain an estimate of the drop in performance expected due to improved deepfake generation. The cross-dataset performance is evaluated using binary real/deepfake labels, since the specific deepfake kinds and engines do not match between the datasets.

Whilst both cross-evaluated models exhibited a drop in performance, the version 1.1-trained model exhibited a more substantial performance decrease when evaluated on version 2.0, with AUROC dropping from 99.7\% to 90.2\%. For real videos the precision was notably low (35.4\%) despite high recall. The opposite is true for fake videos, where the precision was high (98.7\%) but recall was lower (64.2\%). In this configuration, many deepfake videos are classified as real, i.e. the videos in the version 2.0 test set successfully pass as real videos when analysed by the version 1.1-trained model.

The more pronounced degradation in the version 1.1-trained model's performance when tested on version 2.0 is likely due to the higher quality and increased variety of deepfakes in DeepSpeak version 2.0. Consequently, models trained on more diverse and challenging data (e.g., DeepSpeak version 2.0) exhibit superior generalization capabilities, a factor compounded by the inherently less challenging nature of videos in the DeepSpeak version 1.1 test set.

\subsection{Detection of specific kinds of deepfake}

Table~\ref{tab:component_metrics} details the performance of the ResNet-RNN model and its constituent components on the DeepSpeak version 2.0 test set, with different kinds of deepfake treated as separate classes. The ability of the model to correctly identify genuine videos is consistent across the binary and multi-class tasks. The gap in performance between the ResNet model and the ResNet-RNN model is small compared to the gap between the ResNet-RNN model and the FaceNet-RNN model. The clear implication is that the strong overall performance of the ResNet-RNN model is primarily attributable to the features learned from individual frames during the training of the ResNet model, rather than temporal inconsistencies identified using the RNN. The inclusion of the RNN component boosts performance across all classes, increasing the F1-score by between 2\% and 5\%. For the ResNet-RNN model the greatest confusion occurs between lip-sync and avatar videos, two different kinds of deepfake.

\begin{table*}[h]
\centering
\caption{Comparison of different model architectures' performance on DeepSpeak 2.0. We report overall Accuracy and AUROC, alongside per-class Precision, Recall, and F1-Score for 'real' and the different kinds of deepfake classes. The ResNet-RNN model achieves the highest performance across most metrics, demonstrating the benefit of using features from a strong image-based model. All metrics are presented as percentages (\%).}
\label{tab:component_metrics}
% By removing resizebox and using a standard font size command, we can achieve better consistency.
\begin{tabular}{@{}c cc l ccc@{}}
\toprule
 & & & & \multicolumn{3}{c}{\textbf{Per-Class Metrics}} \\
\cmidrule(l){5-7}
\textbf{Model} & \textbf{Accuracy (\%)} & \textbf{AUROC (\%)} & \textbf{Class} & \textbf{Precision (\%)} & \textbf{Recall (\%)} & \textbf{F1-Score (\%)} \\
\midrule

% --- RNN-FaceNet ---
\multirow{4}{*}{RNN-FaceNet} & \multirow{4}{*}{64.95} & \multirow{4}{*}{80.94} & Real & 72.37 & 77.82 & 74.94 \\
& & & Face-swap & 35.36 & 14.77 & 20.23 \\
& & & Lip-sync & 61.94 & 48.28 & 54.05 \\
& & & Avatar & 53.47 & 60.65 & 56.81 \\
\midrule

% --- ResNet ---
\multirow{4}{*}{ResNet} & \multirow{4}{*}{91.09} & \multirow{4}{*}{98.43} & Real & 95.46 & 94.97 & 95.21 \\
& & & Face-swap & 94.88 & 95.97 & 95.42 \\
& & & Lip-sync & 81.64 & 92.31 & 86.65 \\
& & & Avatar & 93.80 & 87.30 & 90.43 \\
\midrule

% --- ResNet-RNN ---
\multirow{4}{*}{ResNet-RNN} & \multirow{4}{*}{94.06} & \multirow{4}{*}{98.81} & Real & 95.61 & 99.72 & 97.62 \\
& & & Face-swap & 97.51 & 96.70 & 97.10 \\
& & & Lip-sync & 92.47 & 90.45 & 91.44 \\
& & & Avatar & 93.41 & 93.36 & 93.39 \\
\bottomrule
\end{tabular}%
\end{table*}

\subsection{Fine-tuning}
\label{sec:finetuing}

\begin{figure}[h]
    \centering
    \includegraphics[width=0.8\textwidth]{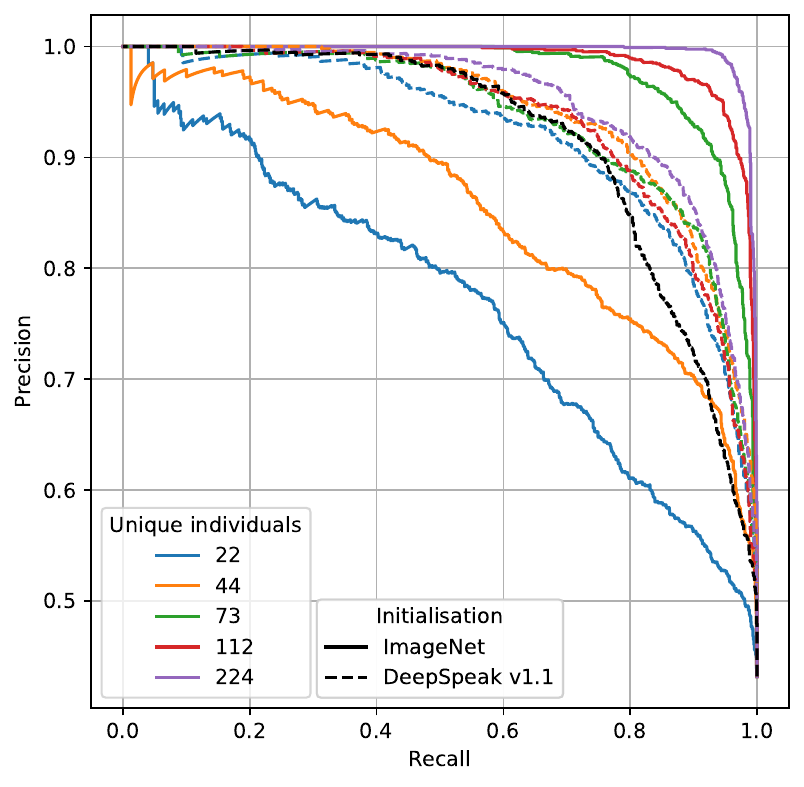}
    \caption{Precision-Recall curves for models fine-tuned using subsets of the DeepSpeak version 2.0 dataset. Solid lines show model performance when trained from an ImageNet initialisation. Dashed lines show performance of models initialised using weights learned from the DeepSpeak version 1.1 dataset. Line colours indicate the number of unique individuals used in each training run. The dashed black line shows the performance of a model trained on DeepSpeak version v1.1 without any fine-tuning.}
    \label{fig:pr_curves_finetuned}
\end{figure}

We assess the utility of a model trained on less recent deepfakes to identify deepfakes generated using novel techniques. Fig. ~\ref{fig:pr_curves_finetuned} shows precision-recall curves for eight models. The four solid lines show precision-recall curves for ResNet models trained using an ImageNet initialisation, using 10\%, 20\%, 50\% and 100\% of the DeepSpeak version 2.0 training dataset for fine-tuning. Dashed lines show precision-recall curves for models trained using the same number of individuals, but initialised using weights obtain by training on the full DeepSpeak version 1.1 dataset. The ImageNet-initialised models are capable of achieving a higher level of overall performance, but only with sufficient data, approximately 100 unique individuals. In cases with substantially fewer videos of individuals, fine-tuning models trained on earlier deepfake datasets offers better performance. We discuss the possible causes and implications of this in section~\ref{sec:disc_ids}.

%%%%%%%%%%%%%%%%%%%%%%%%%%%%%%%%%%%%%%%%%%%%%%%%%%%%%%%%%%%%%%%%%%%%%%%%%%%%%%%%

\section{DISCUSSION}
\label{sec:discussion}

\subsection{Contribution of model components to overall performance}
Almost all model performance is obtained from the features extracted by the ResNet model. Incorporating an RNN into the workflow only raises accuracy by a few percentage points. The same RNN trained using standard FaceNet features performs significantly worse. The clear implication is that deepfake-specific features in individual frames are the key driver of our deepfake detection method, with multi-frame consistency only contributing a small amount to overall performance, even for cutting-edge deepfakes. A focus on learning robust features which can be used for supervised learning would likely yield useful outcomes for the community.

\subsection{Effect of number of unique identities}
\label{sec:disc_ids}
Obtaining numerical estimates of the required dataset size for reliable detection of new-generation deepfakes is essential for being able to rapidly respond to innovative advances in deepfake generation models. These estimates are particularly salient for cases where deepfakes are collected from the wild, but the generation model is unavailable to the wider community. In this case, collection of the dataset represents a bottleneck in response time.

Fig. ~\ref{fig:pr_curves_finetuned} shows the relationship between the number of individuals in the training set, and the performance on the test set for DeepSpeak version 2.0. A clear takeaway is that models trained on previous deepfakes can (still) serve as starting points for fine-tuning in scenarios where only a few dozen new-generation deepfakes are available. However, in a scenario where videos of hundreds of unique individuals are available, ImageNet weights serve as a better initialisation for fine-tuning than weights obtained by training on previous-generation deepfakes. We hypothesise that this can be explained by the nature of deepfake-specific features in versions 1.1 and 2.0 of the DeepSpeak dataset. Whilst there is a degree of commonality between them, many of the version 2.0 deepfakes contain features not present in the version 1.1 dataset. Learning these features is easier for models which begin from more general ImageNet features, rather than more specialised weights which describe features unique to the earlier dataset.

\subsection{Even a binary model learns engine-specific patterns}

\begin{figure}[H] % figure* makes it span two columns
    \centering
    \includegraphics[width=0.8\textwidth]{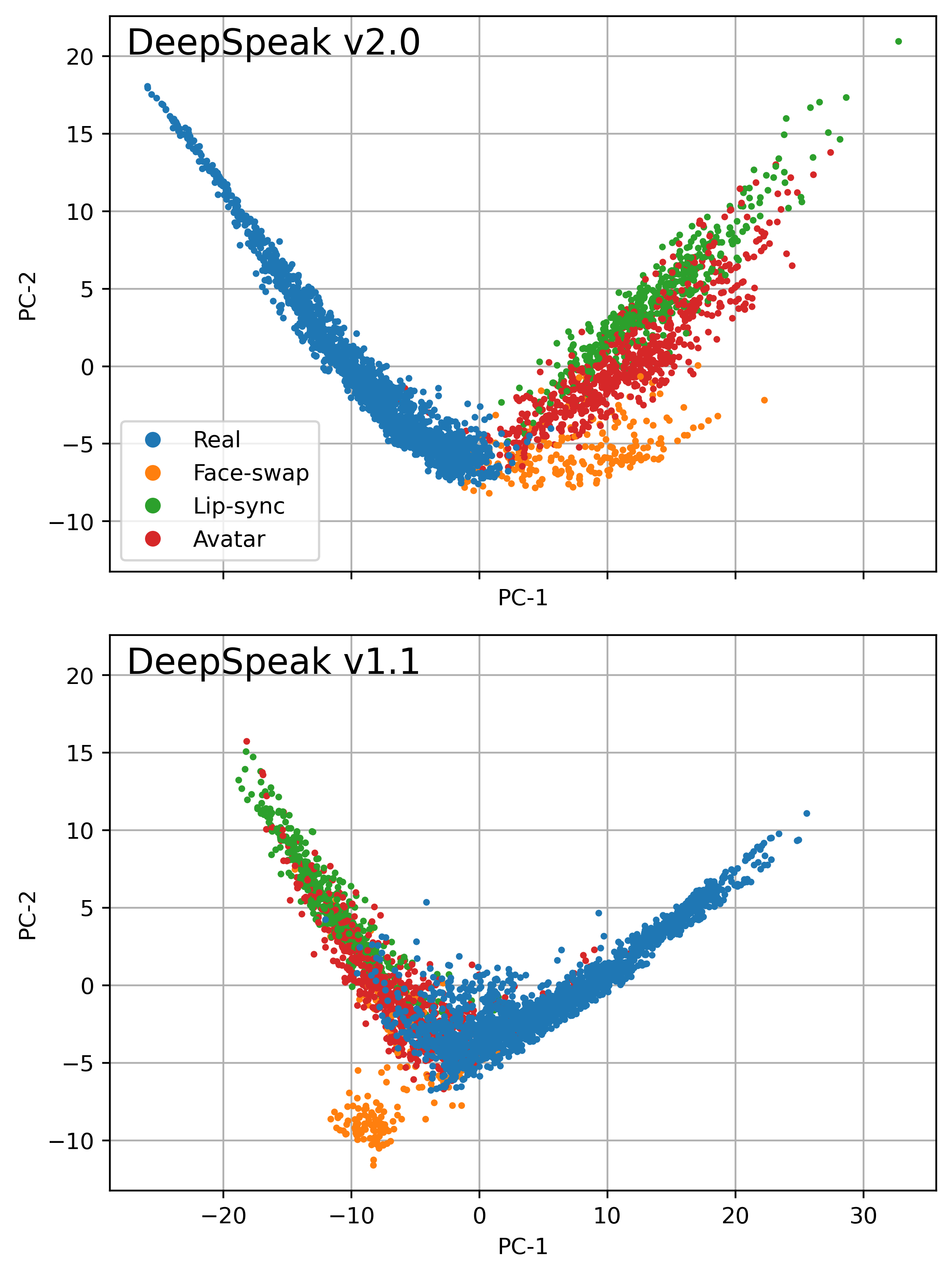}
    \caption{Principal Component Analysis (PCA) plots showing the feature-space representation of the DeepSpeak version 2.0 test set computed using (i) a ResNet model trained on the DeepSpeak version 2.0 train set (top panel) and (ii) a model trained on the DeepSpeak version 1.1 dataset (bottom panel). Both models were trained using binary real/fake labels, and exhibit varying abilities to separate different kinds of deepfake, represented by the different colours of points.}
    \label{fig:pca_multiclass}
\end{figure}

Fig. ~\ref{fig:pca_multiclass} shows the first two principal components of ResNet features extracted from
the DeepSpeak version 2.0 test set. The top panel shows features extracted from a model trained using the DeepSpeak version 2.0 train set, using a binary real-fake criterion. Even though the training procedure does not encourage the model to discriminate between kinds of deepfake, Fig. ~\ref{fig:pca_multiclass} provides evidence that a binary-trained model implicitly learns such distinctions. 

The bottom panel of Fig. ~\ref{fig:pca_multiclass} shows features of the same test set, this time extracted from a model trained only on DeepSpeak version 1.1 data. Of particular interest are the \textit{avatar} deepfakes, as this kind of deepfake is not presented in the version 1.1 dataset. We see that this kind of deepfake are represented as a mixture between a real video and a \textit{face-swap} video. This provides weak evidence that models trained using older data will not learn implicit separations in the same way as models trained using newer data, i.e. exposure to different kinds of deepfake is essential, even for binary classification. To quantify this effect we compare an ImageNet-initialised and version 1.1-initialised model, both fine-tuned on 50\% of the DeepSpeak version 2.0 dataset. We find that the F1 score for the \textit{avatar} class of deepfake, which is not present in DeepSpeak version 1.1, is significantly higher when ImageNet weights are used as an initialisation. Using DeepSpeak version 1.1-initialised weights causes a drop in F1 score from 81.2\% to 61.7\%, a much more significant drop than observed in the other kinds of deepfake.

\subsection{Limitations}
The work presented in this paper has several limitations. Firstly, we have not investigated any weight initialisations other than ImageNet and the DeepSpeak version 1.1-trained models presented in this paper. Whilst we have justified this point from a practical perspective, it is highly plausible that face-specific foundation models~\citep{cai2023marlinmaskedautoencoderfacial} may be more suitable for the transfer learning approach discussed in this paper.

Secondly, we have not yet explored the significance of dataset composition for fine-tuning deepfake detection models. It seems plausible that a small dataset composed of individuals of diverse genders, ages and races will yield better performance than a dataset containing only very similar individuals, however we have not quantified this effect. For all fine-tuning experiments described in this paper, we randomly select subsets of individuals from datasets without reference to characteristics such as gender. It is thus possible that in some cases, for example with very small subsets, that models were trained on datasets which, by chance, contained a low diversity of individuals.

Thirdly, we have not incorporated the available audio data into our methods, despite this being a cutting-edge deepfake detection approach. This point is partially rebutted by the strong performance of our models without the inclusion of audio data. Furthermore, older deepfakes exhibit greater discrepancies between video and audio streams than newer deepfakes. Incorporating the audio stream would therefore serve to increase the gap in detectability between older and newer deepfakes. In this sense, the results presented in this paper are most relevant for assessing when the gap between older and newer deepfakes is incremental, and performance degradation is due to more subtle features.

Finally, and most significantly, we assume that deepfake videos will continue to contain  machine-learnable features which reliably distinguish them from genuine videos. As the capabilities of generative AI continue to advance rapidly, this assumption may well break down. In such a scenario, watermarking and other provenance tracking methods will offer the only recourse for maintaining trust in digital media.
%%%%%%%%%%%%%%%%%%%%%%%%%%%%%%%%%%%%%%%%%%%%%%%%%%%%%%%%%%%%%%%%%%%%%%%%%%%%%%%%

\section{CONCLUSION}
\label{sec:conclusion}
We have demonstrated that a two-stage method, consisting of feature extraction with a simple CNN followed by temporal feature learning using an RNN, is able to reliably discriminate between genuine videos and multiple kinds of cutting edge deepfake, achieving AUROCs of 99.7\% and 99.8\% on versions 1.1 and 2.0 of the DeepSpeak dataset respectively. We have quantified the drop in performance attributable to the advancement of deepfake technology. We observe a drop in the recall of deepfakes of over 30\%. Taken together, these two points suggest that the future of deepfake detection relies on the rapid assembly of high-quality datasets which reflect the most recent innovations in generative capability

We have also explored the question of how much data is required to achieve a given level performance on cutting-edge deepfakes. We show that in cases where only small quantities of new deepfakes with limited individual diversity are available for training, models trained using previous deepfakes are of value. However, assembling larger datasets, containing at least 100 unique individuals, is of critical importance for achieving the highest levels of detection performance on new deepfakes.

Finally, we have shown that the majority of the predictive power of our model is attributable to frame-level features, rather than the temporal relationships between frames. This result suggests that the focus of model development should be on developing robust deepfake-specific features.

\subsection*{Acknowledgments}
\noindent This work was supported by the Alan Turing Institute’s Defence and National Security programme through a partnership with Dstl.

{\small
\bibliographystyle{IEEEtranN}
\bibliography{bib}
}

\end{document}